# Clinically-Validated Innovative Mobile Application for Assessing Blinking and Eyelid Movements

Gustavo Adolpho Bonesso - 0000-0002-8691-7607[1], Carlos Marcelo Gurjão de Godoy - 0000-0001-8846-3242[1(✉)], Tammy Hentona Osaki - 0000-0002-2307-2141[2], Midori Hentona Osaki - 0000-0001-8574-2879[2], Bárbara Moreira Ribeiro Trindade dos Santos - 0000-0002-7792-5920[2], Juliana Yuka Washiya - 0000-0001-5908-0031[2], and Regina Célia Coelho - 0000-0002-4428-9745[1]

[1] Science and Technology Institute, Federal University of São Paulo, São José dos Campos, SP, Brazil

[2] Department of Ophthalmology and Visual Sciences, Paulista School of Medicine, Federal University of São Paulo, São Paulo, SP, Brazil

Corresponding Author: gurjao.godoy@unifesp.br

**Abstract.** Blinking is a vital physiological process that protects and maintains the health of the ocular surface. Objective assessment of eyelid movements remains challenging due to the complexity, cost, and limited clinical applicability of existing tools. This study presents the Bapp (Blink Application), a mobile application developed using the Flutter framework and integrated with Google ML Kit for on-device, real-time analysis of eyelid movements, and its clinical validation. The validation was performed using 45 videos from patients, whose blinks were manually annotated by an ophthalmology specialist as the ground truth. The Bapp's performance was evaluated using standard metrics, with results demonstrating 98.4% precision, 96.9% recall, and an overall accuracy of 98.3%. These outcomes confirm the reliability of the Bapp as a portable, accessible, and objective tool for monitoring eyelid movements. The application offers a promising alternative to traditional manual blink counting, supporting continuous ocular health monitoring and postoperative evaluation in clinical environments.

## 1 Introduction

Eye blinking is the rapid closing and reopening of the eyelids, a spontaneous and involuntary action that preserves ocular surface health and maintains visual clarity by evenly spreading the tear film, thereby playing an essential role in protecting eye health and supporting visual processing [1]. Unfortunately, changes in blinking may occur in conditions such as dry eye syndrome or eyelid ptosis, or after eyelid surgery, and can potentially harm eye health [2–4].

Analyzing eyelid blinking and movement is crucial for monitoring patients with abnormal eyelid behavior. This analysis involves capturing facial images and using specialized software to recognize and measure eyelid movements[5–8]. Nevertheless, such approaches often involve complex systems that are impractical for clinical settings. Additionally, they tend to be imprecise, especially for patients with abnormal eyelid movements, and fail to capture other critical parameters, such as eyelid amplitude movements [5–8].

Fortunately, cameras, including the built-in cameras of mobile devices, can acquire images of fair quality [5–8], which, in turn, may be suitable for measuring eyelid movements using computational tools such as MATLAB [6]. Alternatively, the user would eventually rely on web-based platforms accessible on mobile devices or desktop computers to conduct these studies [5], thereby improving the applicability of these computational resources. However, despite the potential of these resources, assessing eyelid movements remains challenging due to the complexity, cost, and limited clinical applicability of existing tools.

In this work, we present a clinically-validated mobile application designed to assess eyelid movement by recording eye opening and closing over time. Validation occurred by comparing its performance against a ground truth established through frame-by-frame annotations by an ophthalmology medical specialist on a sample of patient videos. The application utilizes machine learning and computer vision tools. It offers a reliable framework for analyzing eyelid movements, providing a scalable and accessible solution for clinical and research use.

## 2 Related Works

Technological advancements have enhanced the analysis of eyelid blinking and movement, particularly in patient monitoring settings in which more resources are available. Also, techniques such as high-speed imaging, deep learning, and smartphone-collected videos now offer portable, accurate, and non-invasive methods for evaluating blink behavior and its connection to various eye conditions. The following works and methods reflect this evolving context.

High-speed imaging captures rapid movements at very high frame rates, enabling precise slow-motion visualization and detailed analysis of fast-moving dynamic events. When combined with digital image correlation (DIC), high-speed imaging allows accurate measurement of eyelid motion during blinking. Such a method captures detailed kinematic data—such as blink duration, eyelid displacement, and peak velocity—allowing a thorough assessment of both spontaneous and reflex blinks [9].

Intelligent vision measurement systems powered by deep learning can analyze eye openness and provide insights into patients' visual function, especially those with dry eye disease. These systems provide information for assessing blink completeness and deliver consistent, precise measurements, improving clinical evaluations [10, 11].

Using smartphones to capture videos has proven effective for collecting raw data on eyelid movements, enabling comparative analysis of blink dynamics in patients. This approach is cost-effective and accessible, making it suitable for both clinical and research uses, as well as for initial self-assessment [12].

Few mobile applications currently provide an objective analysis of eyelid movements. Among them, DryEyeRhythm and EyeScore were developed to assess blink patterns for diagnosing dry eye. However, the latter is not available for download [13, 14]. There is no comparative validation using these apps against public datasets and clinical data.

A different method for analyzing blinks in videos involves running the application on a cloud server. Patient videos are recorded locally using a camera or smartphone and then uploaded to a web server for processing and analysis of eyelid movement. One project that adopted this method employed the Streamlit platform to host the app [5].

Although these technological advancements offer substantial benefits for tracking eyelid movement, challenges persist in standardizing these techniques across various clinical environments and promoting their widespread adoption. Additional research is necessary to confirm the effectiveness and reliability of these technologies across diverse patient groups, including individuals of different ethnicities and age ranges.

# 3 Proposed Method

The Bapp (Blink Application) used in this study was developed through an iterative process spanning three major versions. In the initial version [15], blink detection relied exclusively on the eye-openness probability provided by Google ML Kit. Although this approach enabled preliminary automated analysis, it was limited in distinguishing subtle or partial blinks. To address these limitations, a subsequent version was conceived, improving the algorithm's robustness by introducing multi-threshold logic and a structured decision pipeline, including pseudocode and refined criteria for blink onset and termination [16]. These refinements decreased false detections caused by probability fluctuations and enhanced temporal stability, making a key step toward a clinically usable solution.

The current, innovative version of the Bapp integrates an additional geometric indicator of eyelid separation — the Eye Aspect Ratio (EAR) — to complement the probability-based signal. EAR provides frame-by-frame information about eyelid separation, making the algorithm more sensitive to low-amplitude and incomplete blinks. This dual-signal approach enhances robustness, especially in clinical recordings where eyelid be-havior may deviate from standard patterns. **Fig. 1** summarizes this methodological evolution and the whole processing pipeline implemented in the present paper. It highlights the transition from the original probability-based method (v0) [15] through the threshold-based refinement (v1) [16] to the present version (v2), which incorporates the EAR-based enhancement.

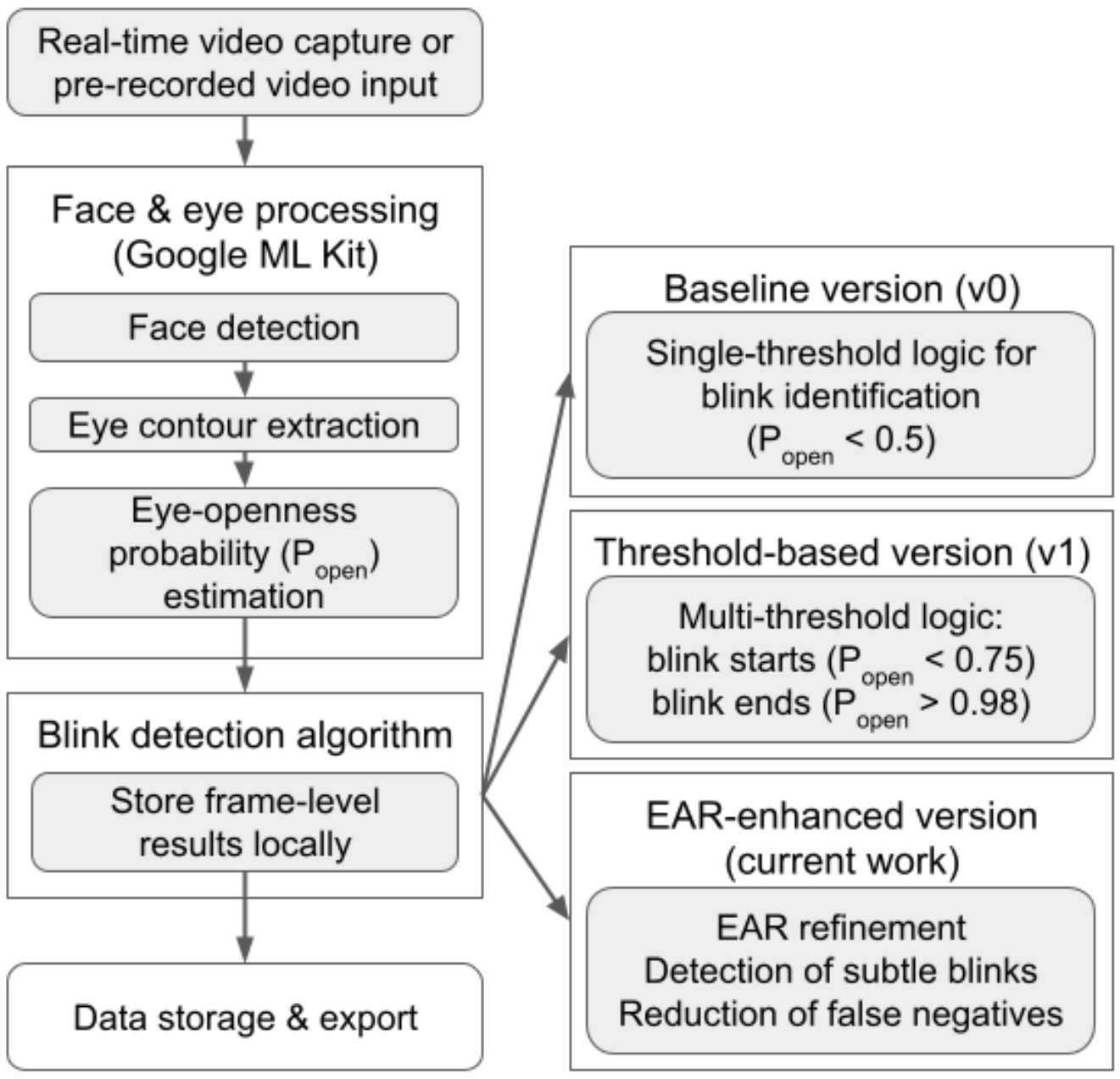

**Fig. 1.** Consolidated flowchart showing the Bapp processing stages: video acquisition, face and eye processing (Google ML Kit), blink detection algorithm, and data export. The diagram also highlights the methodological evolution from the single-threshold approach to the multi-threshold logic and the current EAR-enhanced refinement for detecting subtle and partial blinks. The baseline version (v0) corresponds to the method introduced in [15], and the threshold-based version (v1) corresponds to the workflow presented in [16]. This work introduces EAR-based refinement to improve the detection of subtle and partial blinks.

A structured clinical validation pipeline was implemented after establishing this improved blink-detection framework. The validation relies on annotated videos by an ophthalmology specialist from the Paulista School of Medicine at the Federal University of São Paulo (EPM-UNIFESP. **Fig. 2** illustrates the validation workflow, which begins with video acquisition through Bapp. During use, the application generates a video output that isolates the eye region, enabling an ophthalmology specialist to review the footage without distraction and to precisely annotate blink events.

The specialist performs a frame-by-frame analysis of each output video, registering the start and end frames of every blink in a text file and classifying them as complete or partial. These annotations constitute the ground truth for evaluating the Bapp's performance.

In parallel with the visual output, the Bapp can export an Excel file containing frame-level raw data, including eye-openness probability and EAR values for each eye. These data allow a precise comparison between the signals used by the app and the events annotated by the specialist. With both datasets—annotations and raw data—available, a Python script aligns blink intervals detected by the Bapp with those marked by a specialist, identifying agreements and discrepancies.

This combined workflow, illustrated in **Fig. 2**, ensures that all stages—from algorithm evolution to clinical comparison—are systematically connected. It also provides the necessary information for computing the performance metrics presented in the following sections.

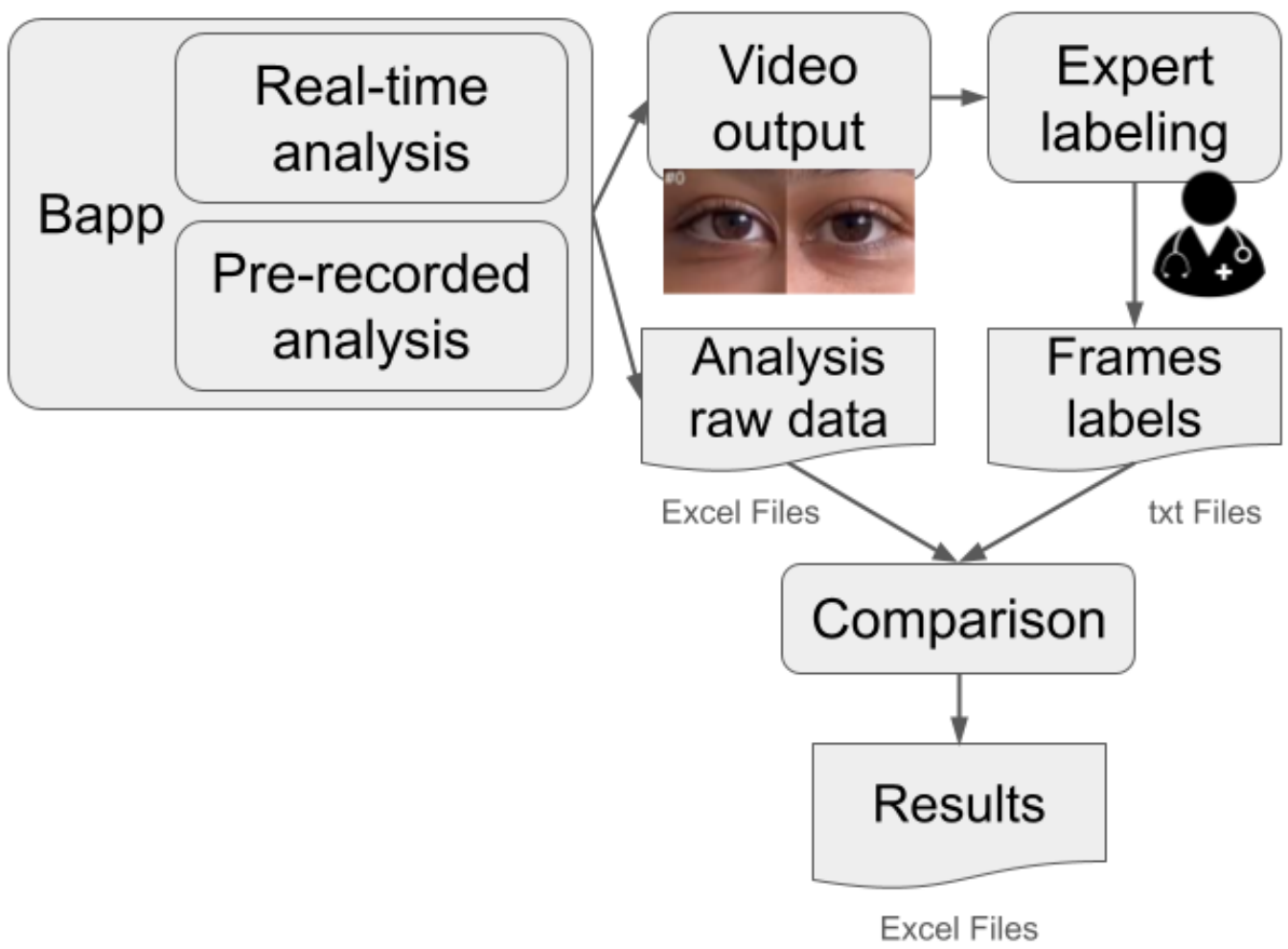

**Fig. 2.** Diagram showing the validation workflow, including the generation of annotated videos, the Bapp analysis (real-time or pre-recorded), and comparison with expert annotations.

Data collection for the clinical validation used the built-in cameras on mobile devices. The app was designed to be cross-platform, reaching nearly all Android and iOS smartphone users. In this study, the iPhone 11 and iPhone 12 (Apple Inc., Cupertino, CA) were used.

### 3.1 Annotated Videos

Spontaneous blinking was recorded bilaterally using the Bapp application, following previously described methods[15, 16]. All recordings were conducted with participants maintaining a primary gaze position under standardized conditions. Individuals with any eyelid, ocular surface, or neurological disorders that could influence blinking were excluded from the study. This study is part of a broader project previously approved by the UNIFESP Ethics Committee (CEP/UNIFESP) under the number CAAE 80417524.2.0000.5505. All participants provided written informed consent prior to their participation in the study.

The application was validated using recordings from 28 healthy adult participants, including 11 men and 17 women, aged 21 to 55 years. As several participants were evaluated more than once, the final dataset comprised 45 video samples: 25 pre-recorded and 20 collected in real time using the Bapp application. All individuals were free from ocular surface disease, neurological disorders, or eyelid abnormalities that could affect normal blink dynamics. This approach ensured that the dataset reflected physiological eyelid behavior under healthy baseline conditions and provided an appropriate reference for assessing the Bapp's performance.

The EPM-UNIFESP specialist annotates blink occurrences in the Bapp-generated output video within a text file. The footage is manually reviewed frame by frame, and the frames indicating the start and end of each blink are recorded. These annotations constitute the ground-truth data, which are then compared with the Bapp output.

**Fig. 3** shows representative frame sequences from video #7 illustrating blink events: (a) depicts a complete blink, whereas (b) depicts an incomplete (partial) blink. Both events are considered blinks.

### 3.2 Eye Blink Detection Algorithm

The interpretation of raw data from Google ML Kit detects the frames where a blink happens. A blink starts when the eye-openness probability drops below 0.75. The blink ends when the eye-openness probability rises above 0.98.

Another method for detecting blinks relies on the EAR, which is much more sensitive. In the proposed approach, eye-openness probability serves as the primary signal for identifying complete blinks. Simultaneously, the EAR is normalized relative to its baseline open-eye value and used as a supplementary geometric indicator. A blink onset is triggered when the normalized EAR drops below 50% of its baseline, signaling a significant transient reduction in eyelid separation. This EAR-based criterion allows the detection of subtle or incomplete blinks that may not produce a significant enough change in eye-openness probability. A blink event is confirmed when either the probability thresholds are met or the EAR criterion is satisfied. **Fig. 4** shows a scenario where the EPM-UNIFESP specialist annotated a blink, but the eye-openness probability failed to detect it. **Fig. 5** displays the frame sequence corresponding to the chart in **Fig. 4**. Using EAR reduces false negatives, especially for subtle blinks.

### 3.3 Blink Raw Data

The Bapp mobile application includes an export function that saves raw eye-openness probabilities and the Eye Aspect Ratio (EAR) for both eyes to an Excel file [16]. **Table 1** displays data from frames 30-45 of video #7, extracted from the Excel file generated by Bapp. The scale ranges from 0 to 1, where 0 indicates a fully closed eye and 1 indicates a fully open eye.

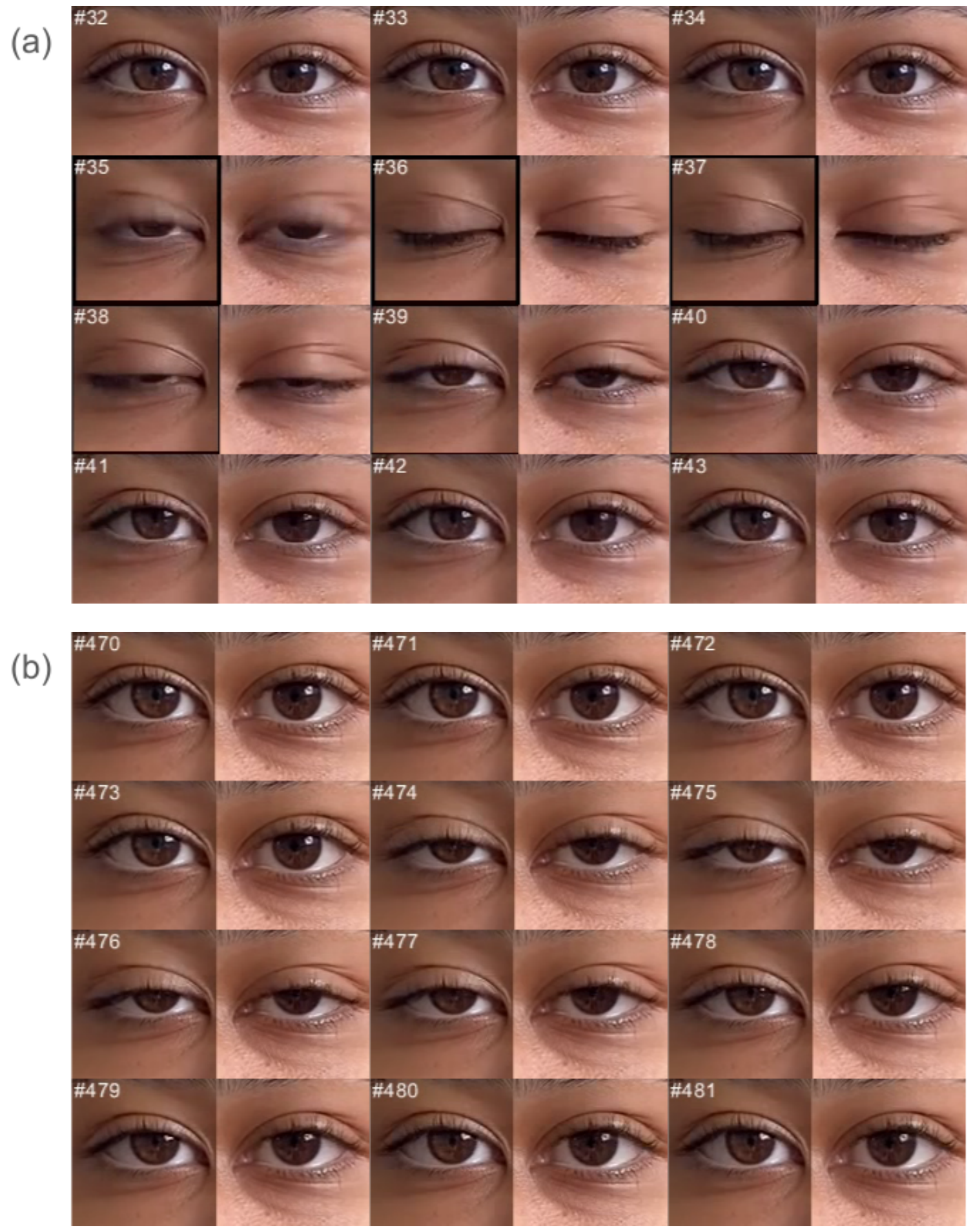

**Fig. 3.** Representative frame sequences illustrating blink events. (a) Complete blink, with full eyelid closure; (b) incomplete blink, in which eyelid closure does not reach complete occlusion.

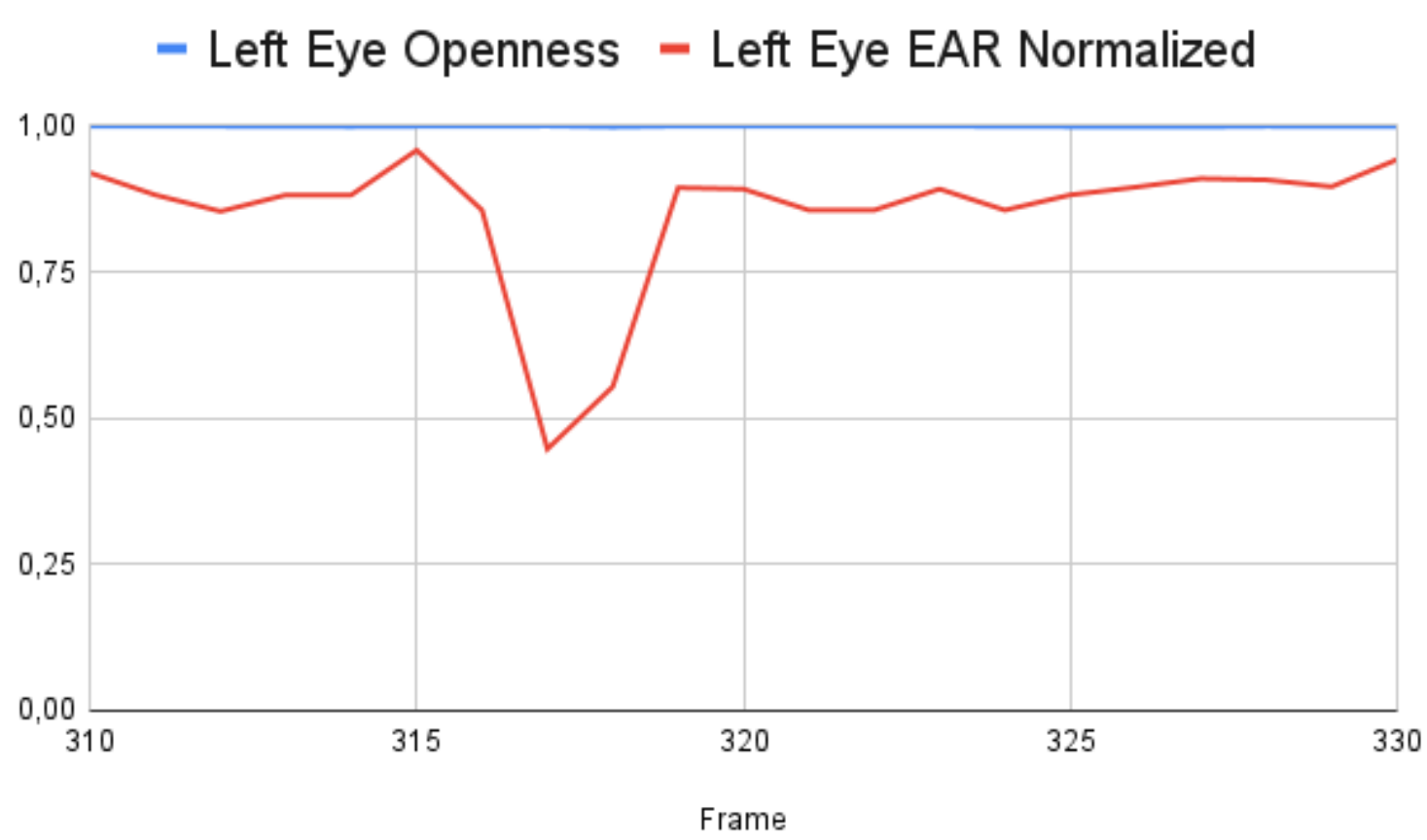

**Fig. 4.** Illustration of eye-openness probability and eye aspect ratio (EAR) normalized for a subtle blink event.

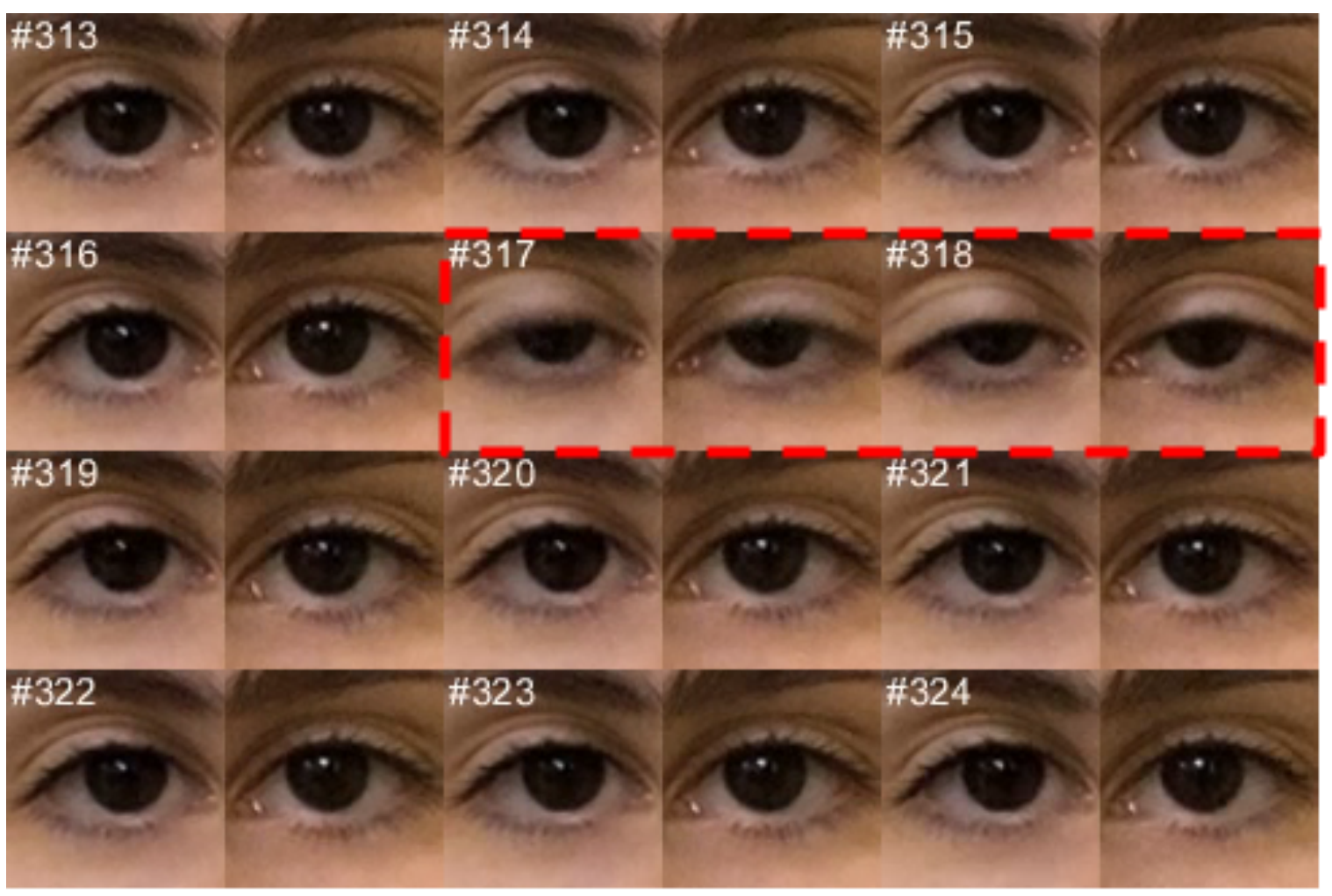

**Fig. 5.** Frame sequence for a subtle blink event.

**Table 1.** Video #7 raw blink data from Excel.

| Frame | Right Eye Openness | Left Eye Openness | Right Eye EAR | Left Eye EAR |
|---|---|---|---|---|
| 30 | 0.992 | 0.999 | 0.326 | 0.326 |
| 31 | 0.988 | 0.998 | 0.326 | 0.326 |
| 32 | 0.992 | 0.998 | 0.326 | 0.326 |
| 33 | 0.993 | 0.999 | 0.337 | 0.326 |

| 34 | 0.992 | 0.998 | 0.326 | 0.322 |
|---|---|---|---|---|
| 35 | 0.993 | 0.691 | 0.236 | 0.225 |
| 36 | 0.024 | 0.003 | 0.191 | 0.188 |
| 37 | 0.029 | 0.001 | 0.189 | 0.200 |
| 38 | 0.084 | 0.019 | 0.229 | 0.222 |
| 39 | 0.994 | 0.987 | 0.275 | 0.255 |
| 40 | 0.994 | 0.996 | 0.288 | 0.296 |
| 41 | 0.995 | 0.998 | 0.324 | 0.304 |
| 42 | 0.991 | 0.999 | 0.339 | 0.313 |
| 43 | 0.993 | 0.998 | 0.335 | 0.328 |
| 44 | 0.993 | 0.997 | 0.332 | 0.328 |
| 45 | 0.993 | 0.997 | 0.339 | 0.326 |

### 3.4 Validation

The raw data exported from the Bapp using the Export to Excel feature were compared, for each analysis, with the annotations provided by the EPM-UNIFESP specialist. A Python script performs this validation by comparing the frame ranges in which the Bapp detects blinks with those identified by the specialist.

To accomplish the statistical calculations, the following information is generated:

- True Positive (TP): When the Bapp detects a blink and the annotations have an overlapping blink in the same frame range.
- False Positive (FP): When the Bapp detects a blink, and the annotations don't have an overlapping blink in the same frame range.
- False Negative (FN): When an annotated blink frame range doesn't have an overlapping blink detected by Bapp.
- True Negative (TN): When an open eyes range is detected by the Bapp and no blink is annotated by the EPM-UNIFESP specialist.

**Fig. 6** shows a confusion matrix.

To calculate statistics from the confusion matrix data, we used the following equations:

$$Precision = \frac{TP}{TP+FP} \quad (1)$$

$$Recall(TP\ rate) = \frac{TP}{TP+FN} \quad (2)$$

$$F1-Score = 2\ x\ \frac{Precision\ x\ Recall}{Precision+Recall} \quad (3)$$

$$Accuracy = \frac{TP+TN}{TP+TN+FP+FN} \tag{4}$$

Precision measures the proportion of correct detections among all detected events. This measurement is especially relevant when the cost of a false positive—detecting a blink that did not actually occur—is high. In this context, a high-precision value indicates that the system produces very few false detections, ensuring that most identified blinks are real.

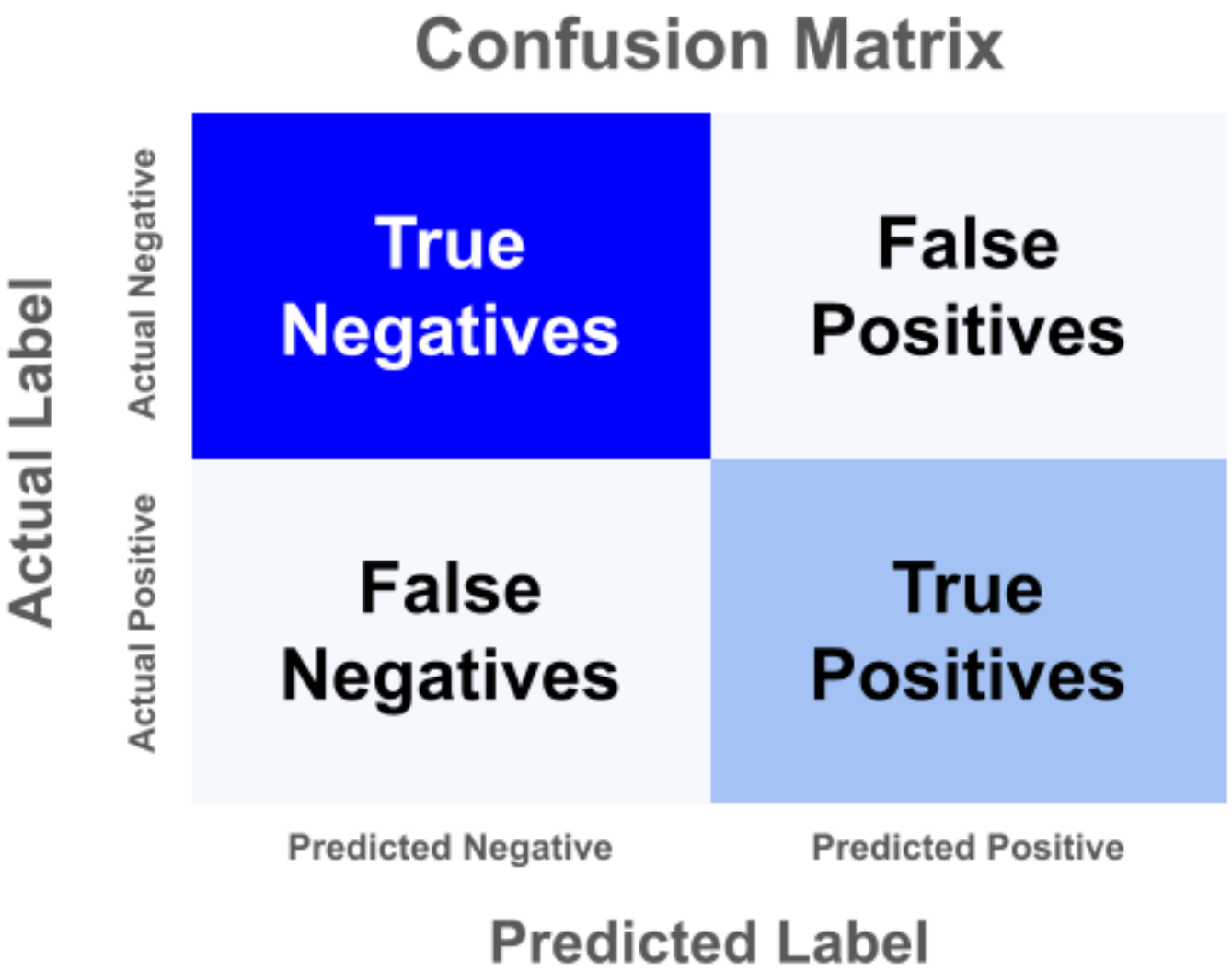

**Fig. 6.** Confusion matrix.

Recall measures how well the system detects all actual events, specifically genuine blinks. A high recall shows that the model finds most actual blinks, reducing missed detections. This metric is fundamental when it is unacceptable to miss any blink.

F1-Score combines precision and recall into a single balanced measure. It provides a more comprehensive assessment of the system's performance, especially when there is a trade-off between maximizing the number of detected blinks (high recall) and ensuring accurate detections (high precision).

Accuracy reflects the overall proportion of correct classifications, including both blinks and non-blinks. Although this value is high, accuracy alone can be misleading when the data are imbalanced—for instance, when there are significantly more non-blink frames than blink frames. Thus, even though it is helpful as a general indicator, it should be interpreted alongside other metrics.

## 4 Results

We used 45 raw analysis datasets from 28 different individuals. The EPM-UNIFESP specialist annotated these 45 analysis-generated videos, totaling 28,975 individual frames. **Fig. 7** illustrates the resulting confusion matrix, demonstrating consistently high accuracy for the derived performance metrics.

The metrics used were:

- Accuracy: 0.98361 (IC95% Wilson: 0.98208 – 0.98501)
- Precision: 0.98407 (IC95% Wilson: 0.98145 – 0.98633)
- Recall (Sensitivity/TP Rate): 0.96967 (IC95% Wilson: 0.96618 – 0.97281)
- F1-Score: 0.97682 (IC95% Bootstrap: 0.97474 – 0.97886)

These results indicate that the system is highly reliable both in correctly identifying true blink events (high recall) and in minimizing incorrect detections (high precision). To ensure statistical rigor, 95% confidence intervals (CIs) were calculated: the Wilson score interval was used for precision, recall, and accuracy, and bootstrap sampling (20,000 resamples) was used to estimate the CI for the F1-score.

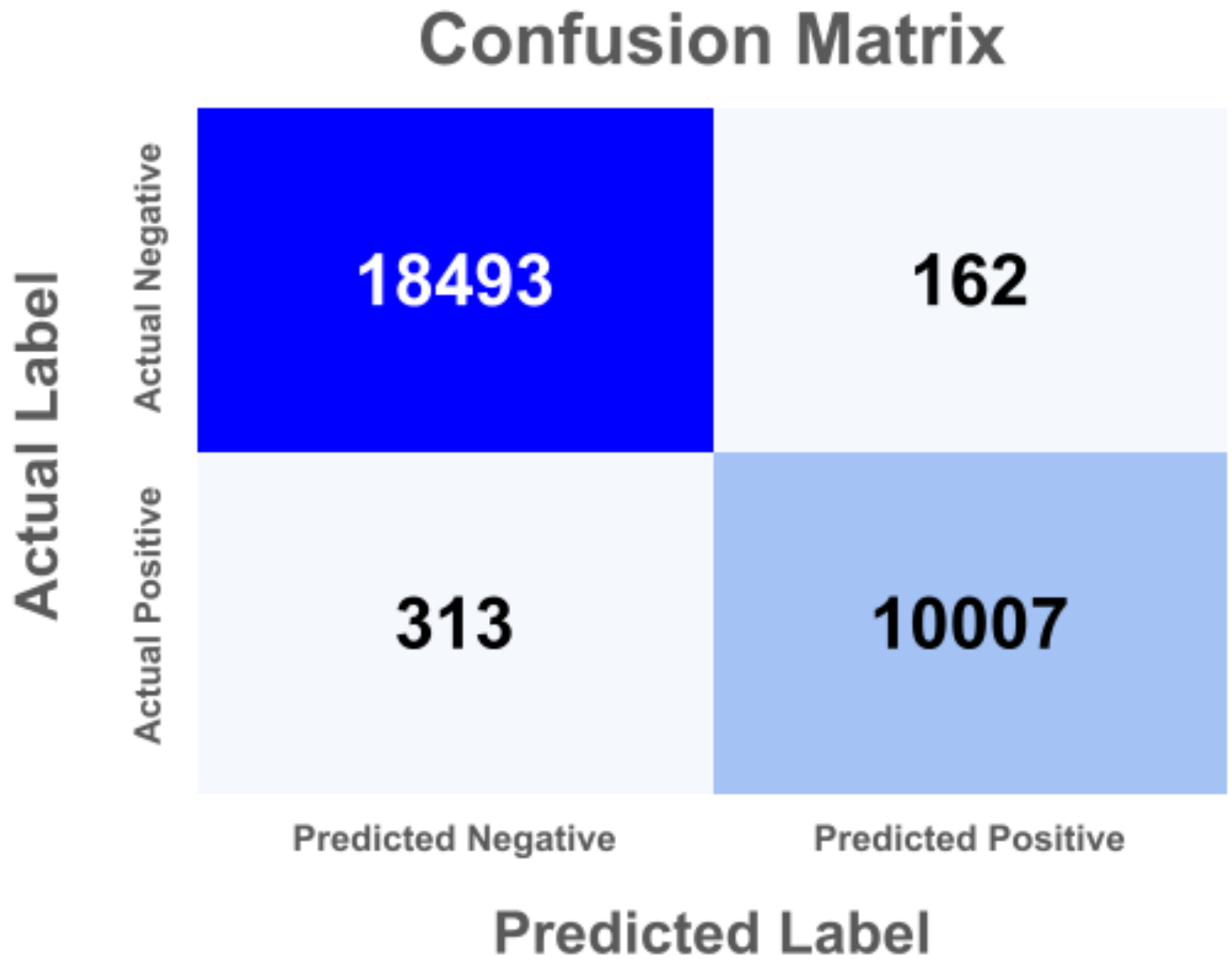

**Fig. 7.** Confusion matrix using the complementary EAR algorithm.

Representative examples from the dataset were analyzed to better understand how these numerical results relate to real-world behavior. Several cases demonstrated the robustness of the dual-signal approach, which combines eye-openness probability with EAR. In video #11, for instance, despite poor lighting, low contrast, and visible noise, both indicators showed a clear and synchronized drop corresponding to the blink annotated by the specialist. **Fig. 8** shows the sequence of frames around the blink event. Although the eyelid outlines are less defined and the overall image quality is poor, the eye-openness probability signal shows a clear dip at frame #22, consistent with the ground-truth annotation. **Fig. 9** presents the related chart combining eye-openness probability and EAR. Both indicators correctly identify the blink, confirming that the algorithm remains reliable even when landmark detection is more challenging due to imaging artifacts.

This example emphasizes the resilience of the Bapp's dual-signal approach (eye-openness probability + EAR), showing that the system can maintain high performance in non-ideal recording conditions—an essential requirement in real clinical settings where lighting and device quality can vary.

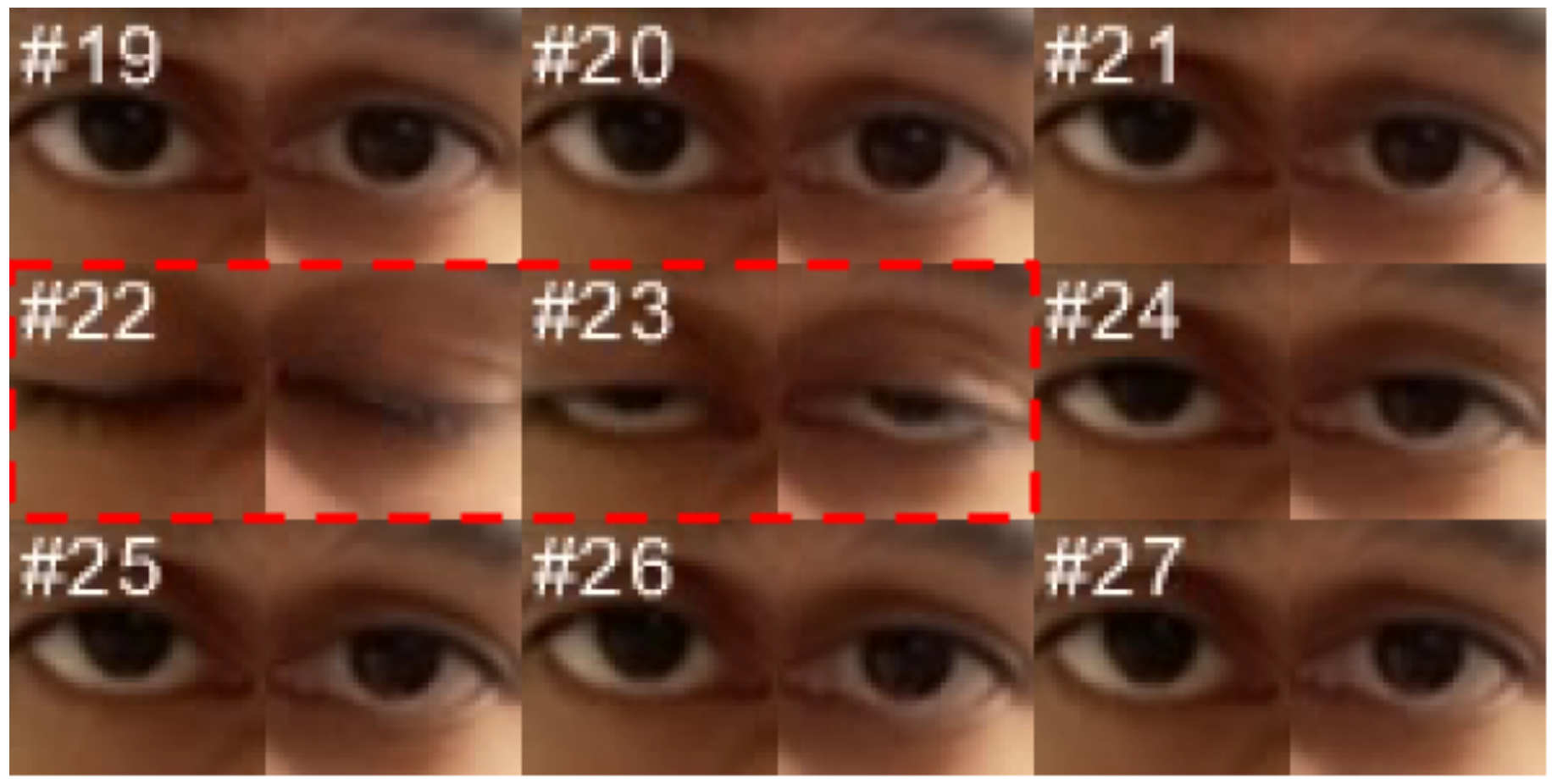

**Fig. 8.** Frame sequence from video #11 showing a blink under poor imaging conditions. Despite the low resolution, uneven illumination, and visible noise in the recording, the Bapp correctly detected the blink at frame #22.

Specific instances were examined to identify the algorithm's limitations. A small number of false-positive events occurred, usually linked to geometric fluctuations rather than misinterpretation of eyelid motion. For example, in video #12, the algorithm detected a blink at frame #6, although the specialist's annotation showed no eyelid closure. **Fig. 10** shows the EAR signals for both eyes. At frame #6, a sharp downward fluctuation crosses the blink-detection threshold, causing a false-positive. However, as shown in the frame sequence in **Fig. 11**, no actual blink occurred. The eye stays open, with no downward eyelid movement. These cases were rare and generally related to low resolution, handheld-camera instability, and slight differences in facial-landmark positioning. Since the EAR is calculated from eye-contour distances, even minor shifts between frames can falsely suggest a reduction in eyelid opening, especially in videos with motion blur or unstable focus. In this case, camera tremor caused a subtle geometric distortion mistaken for eyelid closure. Understanding these cases highlights a known limitation of geometric estimators such as EAR. It underscores the potential for future use of learning-based models, which are less affected by landmark jitter.

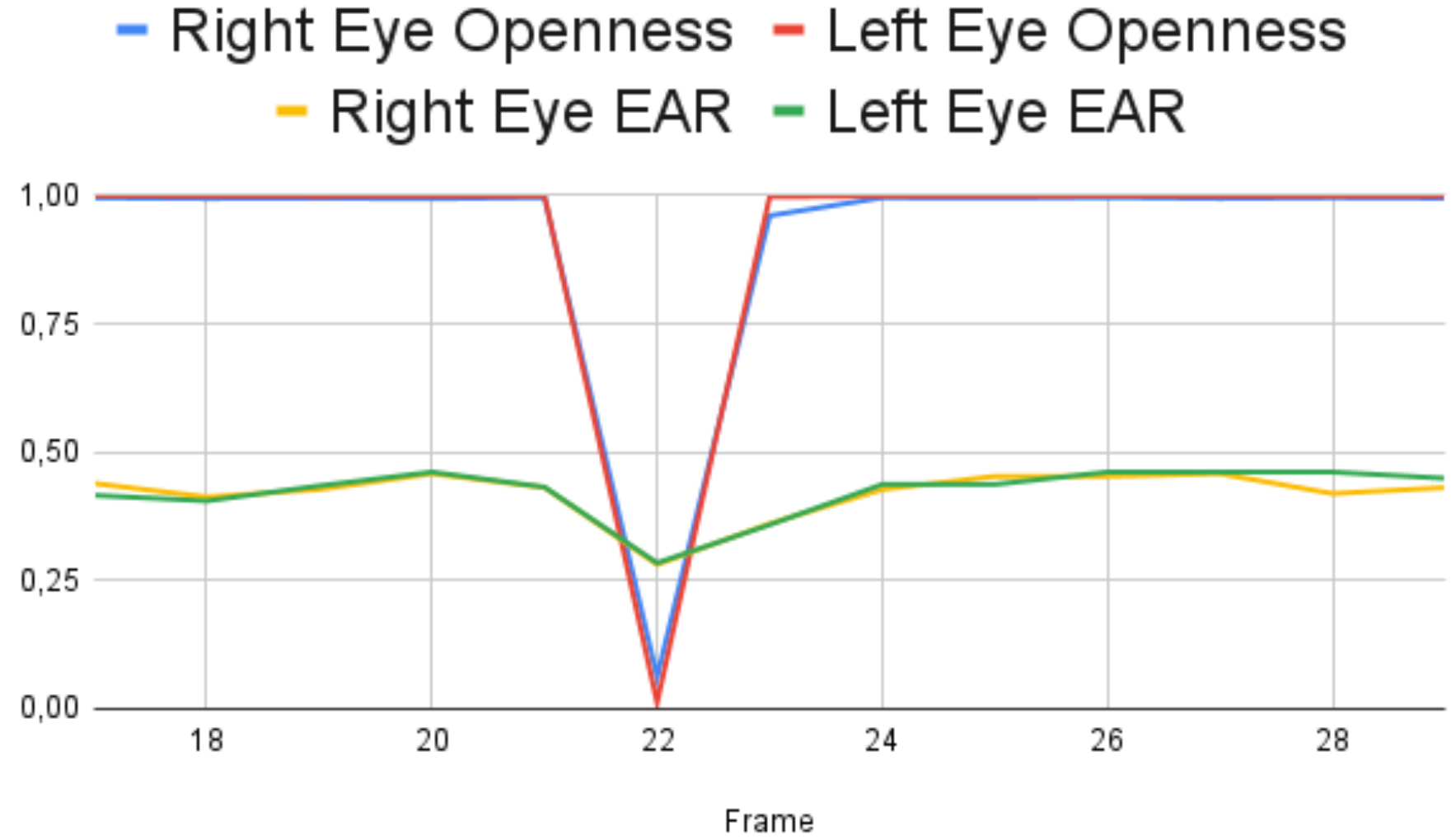

**Fig. 9.** Eye-openness probability and EAR signals for video #11. Both the probability curves and EAR trajectories show a clear drop at frame #22, which aligns with the specialist annotation and confirms correct blink detection even under suboptimal video quality.

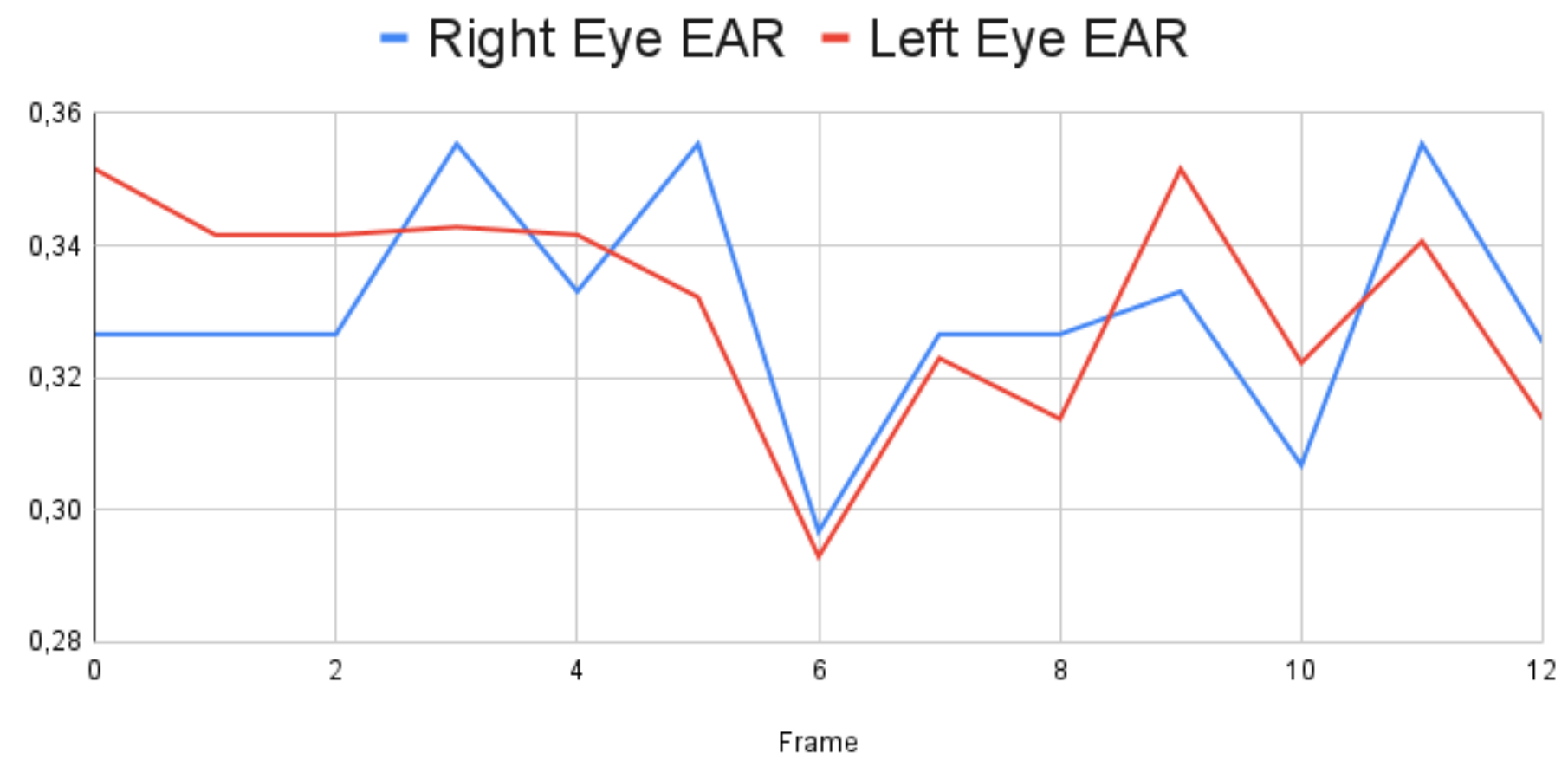

**Fig. 10.** This chart shows the variation of the Eye Aspect Ratio (EAR) for both eyes throughout the recording. A sharp fluctuation at frame #6 triggered a false-positive blink detection, despite no actual eyelid closure. This event illustrates how camera instability and low video resolution can introduce geometric noise, affecting EAR-based measurements.

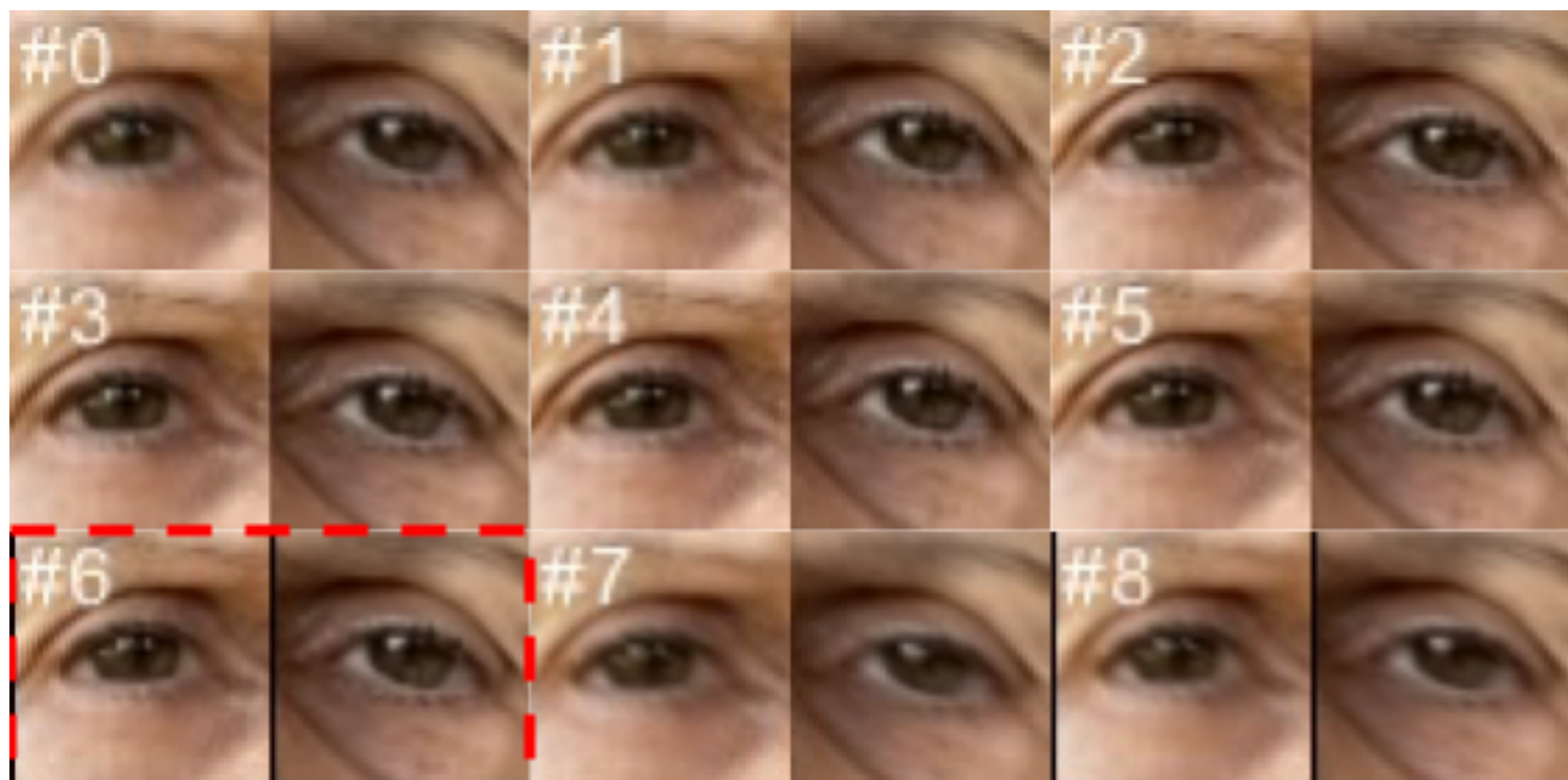

**Fig. 11.** The sequence displays the frames surrounding the false-positive event detected by the EAR algorithm. Although the system flagged frame #6 as a blink, visual inspection confirms that the eyelid remains open, indicating no blink occurred. This example demonstrates that small landmark shifts due to motion and image quality can lead to incorrect EAR-based detections.

Similarly, the system showed a few false-negative detections, usually involving very subtle partial blinks. For instance, in video #1, the specialist annotated a low-amplitude, incomplete blink between frames #131–135. However, neither the eye-opening probability nor the EAR showed enough change to exceed the detection thresholds. **Fig. 12** shows the frame sequence around the annotated interval. The blink is very subtle, with minimal eyelid displacement and no full closure. This slight movement is evident in the raw signals in **Fig. 13**. Both probability curves remain near 1.0, and the EAR values decrease only slightly around frame #133, which is insufficient to meet the algorithm's decision criteria. This example highlights a fundamental limitation of geometric methods such as EAR: very subtle eyelid movements can fall within the natural variation in landmark localization, especially when the contour shift is only a few pixels. Although these cases were rare in the dataset, they emphasize the challenge of detecting micro-blinks or low-amplitude partial blinks with threshold-based methods.

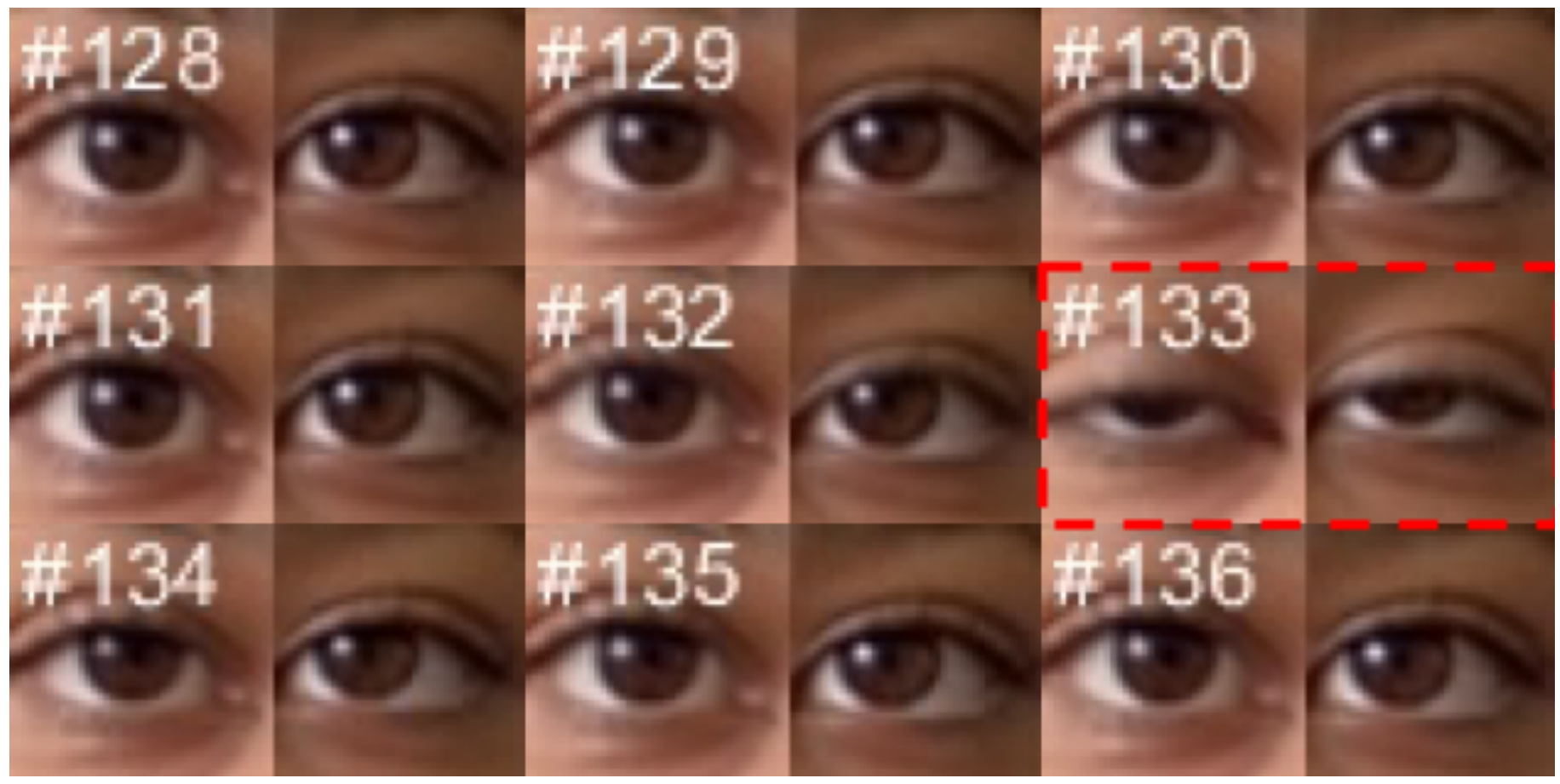

**Fig. 12.** Frame sequence from video #1 illustrating a very subtle partial blink (frames #131–135). Although the specialist correctly identified the blink, the eyelid displacement is minimal, making the event visually subtle and complex for geometric estimators to capture.

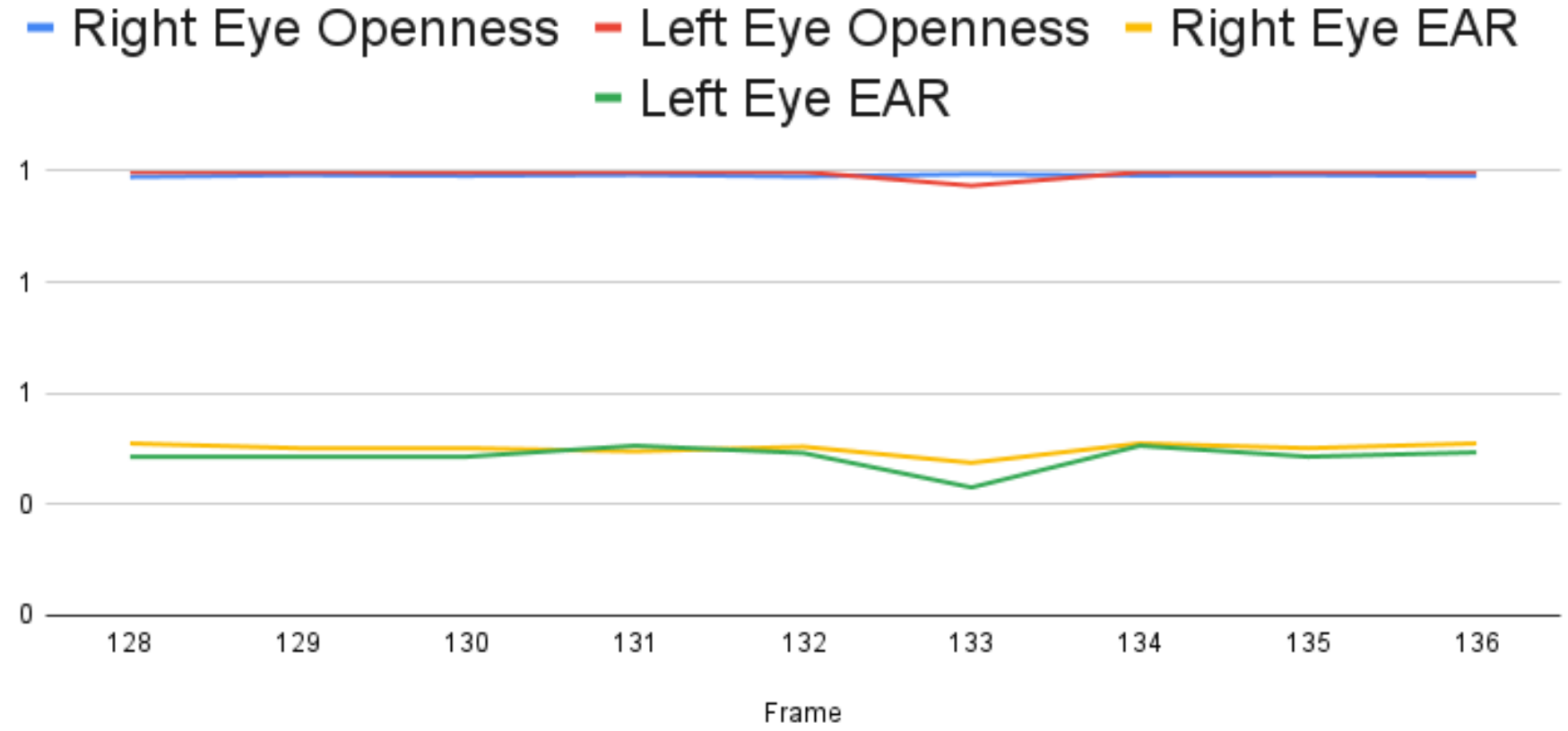

**Fig. 13.** Eye-openness probability and EAR values for video #1. Both signals exhibit only minor fluctuations during the annotated blink interval, falling below the algorithm's detection thresholds and resulting in a false-negative event.

Taken together, the quantitative metrics and qualitative analyses show a consistent pattern: the Bapp performs reliably across a wide range of real-world conditions, whereas the few remaining detection errors can be attributed to well-defined boundary cases such as unstable recordings or very low-amplitude eyelid movements.

## 5 Discussion

The results of this study show that the Bapp, by combining eye-openness probability and EAR, provides reliable, clinically meaningful blink detection in real-world recording environments. As a cross-platform mobile application compatible with both Android and iOS devices, equipped with machine-learning capabilities that analyze eyelid movements in pre-recorded videos and in real time, the Bapp offers broad accessibility and practical utility in clinical and research settings. The system's high precision (98.4%) and recall (96.9%) confirm its ability to accurately detect actual blink events while minimizing false positives, even when videos are captured under diversified lighting conditions, with varying camera stability, or at different camera angles. The precision achieved by the Bapp in previous analyses on public datasets was 95.3% on the Talking Face and 84.3% on the EyeBlink8 dataset [16]. These precision levels enable the Bapp to analyze eyelid movements rather than relying on manual blink counting, thereby facilitating the monitoring of abnormal eyelid activity.

Additionally, the qualitative analysis of individual cases offers valuable insights into the behavior of the current detection framework. As shown in the Results section, the EAR-enhanced algorithm demonstrated both strengths and limitations under challenging conditions. A representative false-positive case occurred in a low-resolution, unstable video, where camera motion caused geometric distortions that briefly made the EAR appear to mimic eyelid closure. In contrast, another example showed that the system remains highly reliable even in poor lighting and noisy conditions, accurately identifying a blink despite low-quality images. A false-negative case illustrated the difficulty of detecting extremely subtle, low-amplitude blinks: the eyelid movement was so slight that neither the eye-openness probability nor the EAR showed a sufficiently distinct change to exceed detection thresholds. These examples highlight the boundary conditions of the probability- and EAR-based approach and help contextualize the quantitative results presented earlier.

A key feature of this study's clinical data collection is the absence of standardization in specific capture parameters, especially lighting conditions and camera angles. While sources suggest optimal results require proper lighting and the subject facing directly toward a stable frontal camera, the present study's validation used 45 videos from patients in environments with diverse illumination. The high performance—98.4% precision and 96.9% recall—despite these variations in real clinical environments—shows the robustness of the Bapp model for analyzing eyelid movement. Using videos with annotations provided by an EPM-UNIFESP specialist in clinical settings enables validation with a more diverse dataset.

The primary limitation of the present work concerns the duration of the analytical process, which depends on multiple factors, including the resolution of the recorded video and the device's computational power. Furthermore, the type of smartphone used for real-time analysis also affects processing performance. Devices with slower processors may experience increased latency or reduced frame consistency, while high-end devices execute the analysis more smoothly. Higher video resolutions require longer processing times, while devices with faster processors can complete the analysis more quickly. However, such devices are generally more expensive than those with slower processors, which take more time to perform the same tasks. Although these differences do not reflect limitations of the algorithm itself, they highlight the dependence of real-time performance on the user's hardware.

Additionally, the original recordings can vary in overall image quality due to differences in smartphone hardware, lens properties, and internal image-processing pipelines. These factors can affect the reliability of eyelid landmark detection and cause fluctuations in the stability of eye-openness and EAR measurements. Modern smartphones use automatic illumination compensation, exposure correction, and noise-reduction algorithms, which may cause frame-to-frame pixel variations and affect the stability of blink-related features. Camera stability also plays a key role; therefore, using a suitable mounting system would mitigate motion artifacts.

Looking forward, several improvements could affect the system's reliability. Using deep learning models that learn spatiotemporal features of eyelid motion may reduce reliance on geometric thresholds and improve the detection of micro- and partial blinks. Additional refinements in stabilization, landmark filtering, or adaptive thresholding could mitigate the effects of camera movement and facial landmark jitter. The platform also presents opportunities to integrate into telemedicine workflows, allowing remote monitoring of patients with ocular surface diseases, facial nerve issues, or post-surgical conditions.

The Bapp's portability, multiplatform availability, and ability to perform both real-time and offline analyses make it a promising tool not only for point-in-time evaluations but also for long-term monitoring. These features enable the application to track blink dynamics over time and assess treatment responses—such as evaluating the effects of botulinum toxin in patients with facial movement disorders or monitoring outcomes in dry eye therapy. Such use cases further emphasize the clinical importance of a reliable, accessible, and automated blink-assessment tool.

To fully establish the Bapp´s generalizability, future work should include validation across larger, more diverse populations spanning a wider range of ages, ethnicities, eyelid shapes, and clinical conditions. Expanding the dataset in this way will be essential to confirm the system's robustness and ensure its suitability for various ophthalmic and neurological contexts.

Overall, the results show that the Bapp is a strong, easy-to-use, and clinically relevant tool for automated blink analysis in many real-world recording scenarios. These features make the mobile application a promising option for long-term monitoring, research, and future clinical use.

## 6 Conclusion

The Bapp application successfully passed clinical validation through annotations performed by an ophthalmology specialist on real patient videos, demonstrating reliable and objective detection of eyelid movements. The system achieved 98.4% precision and 96.9% recall, confirming its high accuracy and robustness in detecting blinks. These results strengthen the Bapp's potential as a portable, accessible, and multilingual tool for continuous patient monitoring and postoperative assessment.

## 7 Disclosure statement

The authors declare that they have no conflict of interest.

## 8 Author contributions

**Gustavo Adolpho Bonesso**: Conceptualization, Methodology, Software, Formal analysis, Investigation, Data curation, Writing – original draft, review & editing.
**Carlos Marcelo Gurjão de Godoy**: Conceptualization, Supervision, Methodology, Writing – review & editing.
**Tammy Hentona Osaki**: Conceptualization, Methodology, Supervision, Validation, Investigation, Resources, Data curation, Writing – review & editing.
**Midori Hentona Osaki**: Validation, Investigation, Resources, Data curation, Writing – review & editing.
**Bárbara Moreira Ribeiro Trindade dos Santos**: Investigation, Data curation, Validation, Writing – review & editing.
**Juliana Yuka Washiya**: Investigation, Data curation, Validation, Writing – review & editing.
**Regina Célia Coelho**: Conceptualization, Supervision, Methodology, Project administration, Writing – review & editing.

## 9 Funding

This work was developed using resources obtained from the São Paulo Research Foundation (FAPESP) [grant numbers 2017/22949-3 and 2012/01505-6].